\documentclass{article}
\usepackage{graphicx}
\usepackage{amsmath}
\usepackage{amssymb}
\usepackage{subcaption} 
\numberwithin{table}{section}
\usepackage{booktabs}
\usepackage[preprint]{corl_2026} 

\title{SA-VLA: State-aware tokenizer for improving Vision-Language-Action Models' performance.}

\author{
  \textbf{Tengyue Jiang}\textsuperscript{2}\quad 
  \textbf{Chunpu Xu}\textsuperscript{3}\quad 
  \textbf{Jiayue Kang}\textsuperscript{4}\quad 
  \textbf{Yao Mu}\textsuperscript{1}\\
  \\[0.2cm]
  \textsuperscript{1}Shanghai Jiao Tong University\\
  \textsuperscript{2}East China University of Science and Technology \quad \\
  \textsuperscript{3}Hong Kong Polytechnic University\\
  \textsuperscript{4}Xi'an University of Electronic Science and Technology \quad 
}

\begin{document}
\maketitle


\begin{abstract}
    Discrete action tokenization provides a compact interface for autoregressive VLA policies, but accurately recovering continuous robot actions from discrete codes remains challenging. Existing tokenizers typically map each discrete code to a fixed continuous action prototype, ignoring the robot’s current proprioceptive state. This limitation is particularly pronounced in manipulation, where the same action token may require different continuous controls under different joint configurations, object poses, and contact conditions. We therefore propose SA-VLA, a state-aware action tokenizer that conditions action decoding on robot state. We study two state-injection mechanisms for VQ-based action tokenization: cross-attention between state and action features, and a lightweight state adapter that predicts action-wise modulation factors for state-conditioned action modulation and reconstruction. The adapter formulation expands the effective support of a finite codebook by allowing each discrete token to represent a family of state-dependent continuous actions, while preserving the efficiency and compatibility of discrete action modeling. Integrated into an LLM-based VLA policy, SA-VLA supports both autoregressive and parallel action-token decoding with minimal changes to the model interface. On 12 RoboTwin manipulation tasks, SA-VLA improves the average success rate from 0.29 to 0.56 over the strongest tokenizer baseline. In zero-shot sim-to-real experiments on three real-world tasks, it further improves average success from 0.15 to 0.33 over the strongest tokenizer baseline. These results demonstrate that state-conditioned action decoding is a simple and effective mechanism for reducing the compression gap in discrete VLA policies.
\end{abstract}



\section{Introduction}

Visual-Language-Action (VLA)\cite{brohan2022rt,kim2025fine,li2025simplevla} models have recently emerged as a highly promising research paradigm in the field of robotic manipulation, capable of directly mapping visual observations and language instructions to action sequences.Generally, VLA\cite{cheang2024gr} models can be divided into two categories based on how they generate actions: one category generates discrete action tokens that are then decoded into continuous actions via an action tokenizer(e.g., OpenVLA\cite{kim2024openvla}, VQ-VLA\cite{wang2025vq})while the other category directly generates continuous actions (e.g., $\pi$0\cite{black2024pi_0}).A common approach for generating discrete action tokens involves mapping visual predictions and language instructions into action tokens through a pretrained VLM, which are then decoded into continuous action values via an action tokenizer. 

However, how to convert the discrete action tokens output by the VLM\cite{bai2025qwen3} into continuous actions remains a significant bottleneck limiting model performance\cite{shiba2026compression}. Previous research has made various attempts in this direction. For example, the binning operation used in OpenVLA\cite{kim2024openvla} is simple and easy to implement but its tokenization granularity is low.The FAST\cite{pertsch2025fast} method offers good compression performance for high-dimensional information; however, the sequence length obtained from compressing actions of the same length may be inconsistent, making it difficult to train. The residual VQ-VAE\cite{van2017neural} approach in VQ-VLA\cite{wang2025vq} is data-driven, yet its compression performance still leaves room for improvement.

To address this, we propose a state-aware action tokenizer, which introduces state information into the action tokenizer. We explore two approaches for incorporating state information. The first introduces state information through a lightweight adapter module, while the second integrates state and action information via cross-attention\cite{vaswani2017attention}. Both approaches achieve significant improvements in both simulation and real-world robot experiments.

Between the two approaches, the method that introduces state information via a lightweight adapter module achieves greater improvements than the cross-attention approach. We attribute this to the fact that traditional VQ-VAE\cite{van2017neural} can only decode a limited set of actions, which constrains model performance. By introducing the adapter module, we enable state-informed prediction of an continuous action space through a regression formulation, thereby enhancing both model performance and action decoding accuracy.

  In summary, our contributions are as follows:
  
1.We propose a state-aware tokenization approach, introducing state information into the action tokenizer.

2.We explore optimal performance through different integration strategies of state information.By introducing a lightweight adapter module to address the limitations of the traditional VQ-VAE\cite{van2017neural}'s finite codebook.

3.Our method achieves strong performance in both simulation and real-world robot experiments


\section{Related works}
\label{sec:Related works}

	\subsection{Vision-Language-Action Models}
Vision-Language-Action Models are considered foundational models in the embodied AI field. They use a pre-trained VLM to extract action tokens from images and language instructions. One approach employs an action tokenizer to convert discrete action tokens into continuous action values, while another uses an action expert that leverages the keys and values from the VLM. Building on this foundational model, recent studies have explored various forms of VLAs, such as those based on discrete diffusion models\cite{liu2026mmada}\cite{liang2025mm}, integrations of world models with VLAs\cite{team2026being},chain-of-thought for action generation\cite{zhao2025cot,liu2026last},and the incorporation of event information into VLAs\cite{zhai2026vla}.
\subsection{Discrete Action Tokenizers}
Action tokenizers are a key component of autoregressive transformer-based VLA models. Recent research has explored various forms of action tokenizers. For example, OpenVLA discretizes actions through uniform binning, where each action value corresponds to a bin, but this tokenization approach is overly coarse. FAST\cite{pertsch2025fast} transforms action values from the time domain to the frequency domain using DCT, followed by BPE to obtain corresponding discrete action tokens. However, due to the nature of BPE, action sequences of the same length may be mapped to token sequences of varying lengths. VQ-VLA uses residual VQ-VAE\cite{van2017neural} to establish a mapping between action values and tokens, but we believe the granularity of this action tokenization remains insufficient. In our work, we propose a state information-guided action tokenizer, which offers stronger generalization capability and positional adaptability.Current tokenizers often borrow tokenizer methods from the field of natural language processing for text or from the field of computer vision for image generation\cite{yang2026vaevq,chen2025softvq,zhu2025addressing}, while neglecting the inherent characteristics of VLA themselves.
	

\section{Methods}
\label{sec:Methods}

\subsection{Problem Formulation}

For a robotic arm manipulation task, given a dataset $\{(o_1, s_1, a_1), \ldots, (o_t, s_t, a_t)\}$ with language instruction $L$, where $o_t$ denotes the observed RGB image, $s_t$ denotes the state of the robotic arm (typically represented by the rotation angle of each joint), and $a_t$ denotes the target action of the robotic arm at the current timestep, our goal is to train a policy $\pi_\theta(a_{t:t+k} \mid o_t, s_t, L)$ that predicts a block of future $k$ actions conditioned on the current observed image, state, and language instruction.

\subsection{State-aware Action Tokenizer}
\begin{figure}[htbp]
    \centering
    \includegraphics[width=1\textwidth]{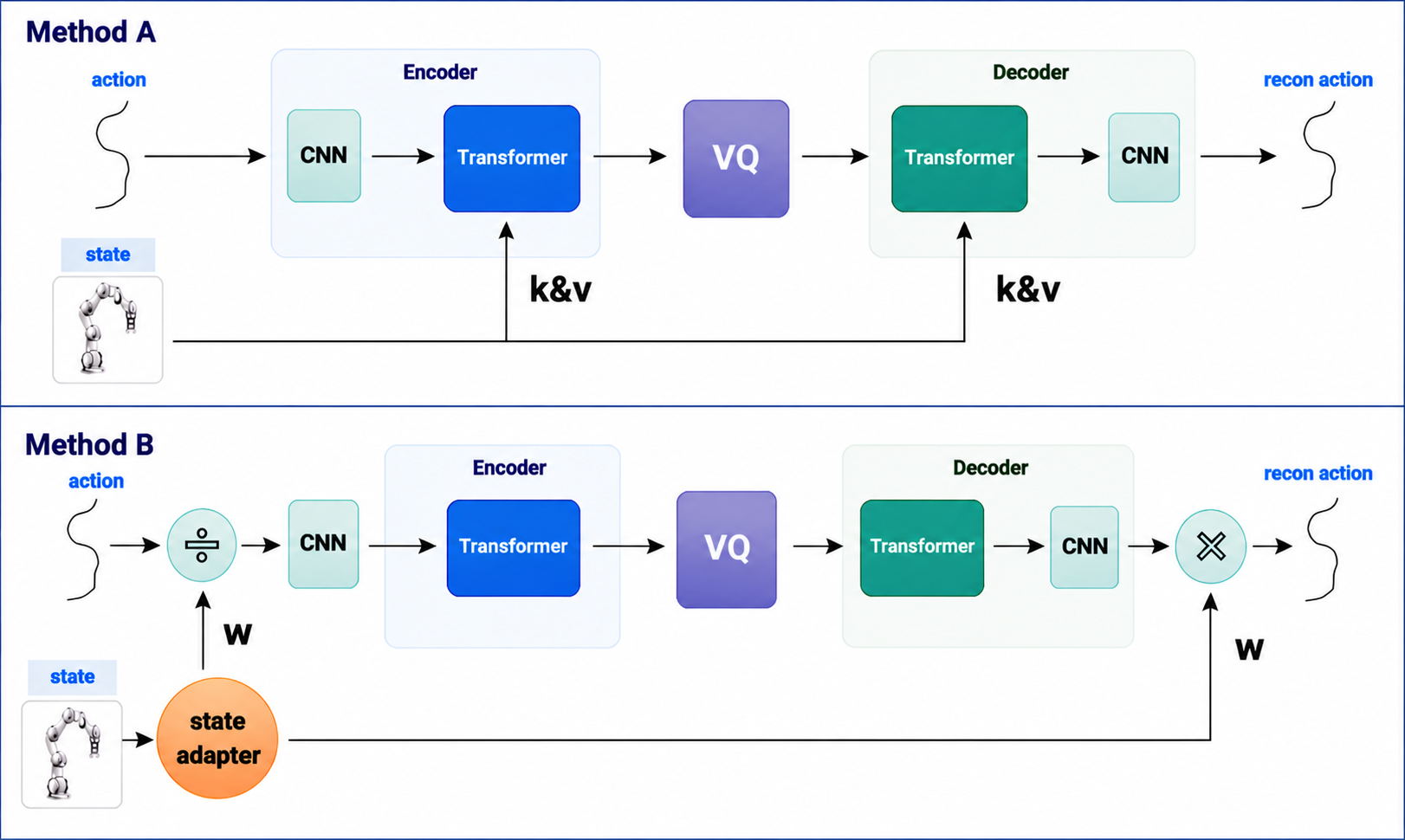}
    \caption{The tokenizer architecture.As shown in the upper part of the figure, Method A integrates state information into the tokenizer by performing cross-attention with the action. As shown in the lower part of the figure, Method B integrates state information into the tokenizer by modulating the action through a lightweight adapter.}
    \label{fig:myfigure}
\end{figure}

First, we briefly introduce the base VQ-VAE\cite{van2017neural}. The base model uses CNN and Transformer\cite{vaswani2017attention} encoders to extract temporal action features. These action features are then quantized using a codebook, and fixed-length discrete action embeddings are output. After that, decoding is performed through a decoder with a symmetric structure to the encoder, outputting the reconstructed actions.Based on this action tokenizer, we consider incorporating state information to improve the accuracy of action decoding. We explore two ways to introduce state information into the action tokenizer. The overall model architecture is illustrated in Figure~\ref{fig:myfigure}.

Our motivation for introducing state into the tokenizer is that when using delta actions, the same delta action may correspond to different physical outcomes in certain situations. For example, when a robotic arm is grasping a bottle, the same delta action for grasping might be identical, but different grasp positions can determine whether the bottle is successfully picked up. If all actions corresponding to different grasp positions are mapped to the same action token, the performance will naturally degrade. This highlights the importance of incorporating state information.Inspired by the interaction between VLM tokens and action experts in $\pi$0\cite{black2024pi_0}, we propose a method that performs cross-attention between state and action within the Transformers of both the encoder and decoder to incorporate state information, denoted as Method A. The other approach is inspired by human behavioral habits: humans often predict future actions based on the current state before acting, for example, when the hand approaches a cup, it is likely to grasp it. Therefore, we propose using a lightweight adapter to map state information into a scaling factor for each dimension of the action sequence, thereby predicting the scale of action generation with the assistance of state information.This way is denoted as Method B.

Below, we describe how the model performs forward propagation in detail. For Method A, given an input action sequence $a$, we first extract its intermediate features using an encoder composed of a CNN and two layers of Transformer\cite{vaswani2017attention}, where the keys and values in the two Transformer layers are both derived from the state input. The intermediate features output by the encoder are denoted as $Z \in \mathbb{R}^{R \times d}$.After the quantization operation of the VQ-VAE\cite{van2017neural} codebook, the resulting quantized vectors are denoted as $\hat{z}_i'$. The quantization process for embedding $\hat{z}_{i}$ with codebook $Z$ can be formulated as follows:

\[
q_{i} = \arg \min \| z_{i} - \hat{z}_{i} \|_2,
\]

where $q_{i}$ is the nearest neighbor lookup index from the codebook $Z$. After the quantization operation, the reconstructed $\hat{a}$ is obtained through a decoder that is symmetric to the encoder. It should be noted that the keys and values in the two Transformer\cite{vaswani2017attention} layers of the decoder also use the raw state data.

Method B is similar to Method A. Given an input action sequence $a$, the state information first passes through a lightweight adapter composed of MLP and Activation functions such as sigmoid\cite{elfwing2018sigmoid} to obtain a scaling factor $w$ for each dimension of $a$, resulting in a transformed action $a_{\text{trans}} = a \div w$. The transformed action $a_{\text{trans}}$ then passes through an encoder with the same structure to extract its intermediate features, where the Transformer\cite{vaswani2017attention} does not incorporate state input. The intermediate features output by the encoder are denoted as $Z \in \mathbb{R}^{R \times d}$. After undergoing the same quantization operation, the quantized features pass through a decoder symmetric to the encoder to obtain the reconstructed transformed action $\hat{a}_{\text{trans}}$. To maintain symmetry, $\hat{a}_{\text{trans}}$ is multiplied by $w$ to obtain the final reconstructed action $\hat{a}$. In this case, the Transformer\cite{vaswani2017attention} does not use state input either.

Following the training paradigm of VQ-VAE\cite{van2017neural}, we define the overall loss $\mathcal{L}$ as a weighted combination of the reconstruction loss $\mathcal{L}_{\text{res}}$, the vector quantization (VQ) loss $\mathcal{L}_{\text{codebook}}$, and the commitment loss $\mathcal{L}_{\text{commit}}$:

\[
\mathcal{L} = \|\hat{a} - a\|_2^2 + \lambda_1 \cdot \| \text{sg}(x) - q(x) \|_2^2 + \lambda_2 \cdot \| x - \text{sg}(q(x)) \|_2^2,
\]

where $\lambda_1$ and $\lambda_2$ are balancing hyperparameters, $\text{sg}(\cdot)$ denotes the stop-gradient operation, $x$ represents the encoder output features, and $q(x)$ denotes the quantized features from the codebook.
\subsection{Integrating  State-aware Action Tokenizer in VLA}
\begin{figure}[htbp]
    \centering
    \includegraphics[width=0.8\textwidth]{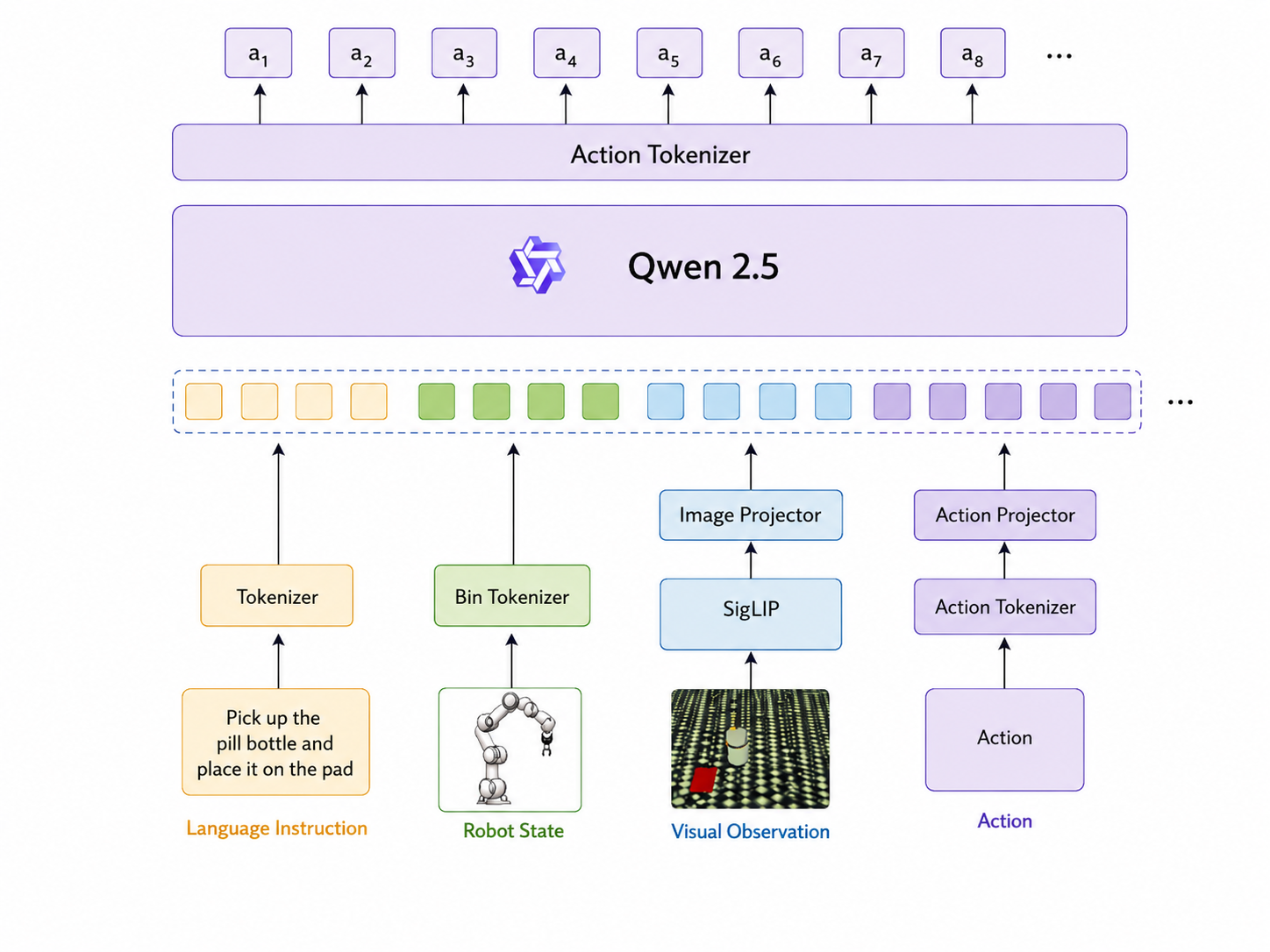}
    \caption{VLA architecture}
    \label{fig:vlafigure}
\end{figure}
The overall model architecture is illustrated in Figure~\ref{fig:vlafigure}.Our goal is to train a VLA\cite{bu2025univla}\cite{cen2025worldvla} policy $\pi_{\theta}(a_{t:t+k} \mid o_t, s_t, L)$ that predicts a block of future actions conditioned on the current observation image $o_t$, robot state $s_t$, and language instruction $L$. For the language description $L$, we use the base LLM's native tokenizer to obtain text tokens. For the joint states $s_t$, we adopt the approach outlined in FAST~\cite{pertsch2025fast} to discretize each joint dimension into 256 bins, yielding a sequence of discrete state tokens. For processing the observation image $o_t$, we resize each image to $224 \times 224$ and apply SigLIP-SO400M-patch14-224~\cite{zhai2023sigmoid} as the image tokenizer, producing a $16 \times 16$ grid of patch\cite{dosovitskiy2020image} embeddings, i.e., 256 continuous image tokens per frame.

For the input action sequence, we use our proposed \textbf{state-aware tokenizer (SA-VLA tokenizer)} to transform continuous action sequences $a'$ into discrete action tokens $\tilde{a}$. Unlike prior tokenizers, our approach introduces state-conditioned decoding via a lightweight adapter that predicts a scaling factor $w$ for each action dimension, transforming the action as $a_{\text{trans}} = a \div w$ before quantization. This effectively reformulates the discrete VQ-VAE\cite{van2017neural} into a regression problem, enabling the generation of a continuous action space while preserving the benefits of a fixed codebook.

To delineate modality boundaries in the input tokens, we introduce special functional tokens: \texttt{t\_bos/t\_eos} for text, \texttt{s\_bos/s\_eos} for state, \texttt{i\_bos/i\_eos} for image, and \texttt{a\_bos/a\_eos} for action. During training, we serialize the context and targets into a single sequence, for example:

{\small \texttt{[t\_bos, t\_tokens, t\_eos, s\_bos, s\_tokens, s\_eos, i\_bos, i\_tokens, i\_eos, a\_bos, action\_tokens, a\_eos]}}

This design yields a consistent tokenized interface across modalities, enabling straightforward autoregressive learning of the action sequence conditioned on language, state, and vision.

We evaluate our state-aware tokenizer under two decoding strategies:

\textbf{Autoregressive Decoding.} In this setting, the VLA generates action tokens sequentially via next-token prediction. The training objective minimizes the negative log-likelihood of predicted action tokens given previous tokens and the full context:

\begin{equation}
\mathcal{L}_{\text{AD}} = -\sum_{i=1}^{h} \sum_{r=1}^{R} \log P(q_{i,r} \mid q_{<}, o_t, s_t, L),
\end{equation}

where $q_{<}$ denotes previous action tokens. 

\textbf{Parallel Decoding.} Our tokenizer produces fixed-length discrete action sequences, enabling parallel decoding\cite{kim2025fine} where the entire action block\cite{zhao2023learning} is generated in a single forward pass. We adopt bidirectional attention and placeholder embeddings for action positions, with the training objective:

\begin{equation}
\mathcal{L}_{\text{PD}} = -\sum_{i=1}^{h} \sum_{r=1}^{R} \log P(q_{i,r} \mid o_t, s_t, L).
\end{equation}

Parallel decoding significantly improves inference efficiency while maintaining high performance. Across both settings, the LLM remains unchanged and continues to output discrete tokens, with only minimal modifications to the input interface and attention mask.

\section{Experiments}
\label{sec:Experiments}

	\subsection{Simulation Experiments}
\subsubsection{Experiment Setup}
Our simulation experiments were conducted on the RoboTwin\cite{chen2025robotwin} platform in clean mode, with 12 tasks, each containing 1,600 trajectories, resulting in a total of 19,200 simulation data entries for training and testing. The specific task types are detailed in Appendix~\ref{app:Dataset Statistics and Task Metrics}. To perform comparative experiments and validate the effectiveness of our approach, we used the same Qwen2.5-based VLM followed by different action tokenizers. The following groups were set up: (1) the bin-based tokenization method represented by OpenVLA\cite{kim2024openvla}; (2) the tokenization method represented by FAST\cite{pertsch2025fast}; (3) the VQ-BET tokenization method represented by VQ-VLA\cite{wang2025vq};(4) Method A as described in our approach; (5) Method B as described in our approach. All groups were trained under the same hyperparameter settings and tested on RoboTwin\cite{chen2025robotwin}.The details regarding training and testing are provided in the Appendix~\ref{app:Detailed Simulation Evaluation Setups}.
\subsubsection{Simulation Results}
We evaluated the models on the same 12 tasks from the RoboTwin benchmark used during training. For each task, we conducted 100 rollouts, where a success was recorded as 1 and a failure as 0, and then computed the success rate for each task. The experimental results are shown in Table~\ref{tab:tokenizer_performance}. The results demonstrate that our methods, Method~A and Method~B, achieve significant improvements over previous action tokenization approaches under both autoregressive and parallel decoding settings. Specifically, Method~B achieves an average success rate of 56\% across the 12 tasks under autoregressive decoding, which is a 23\% improvement over the binning tokenizer, a 40\% improvement over the FAST tokenizer, and a 28\% improvement over the VQ-BET tokenizer.

\begin{table}[htbp]
\centering
\scriptsize
\caption{Tokenizer performance comparison on RoboTwin}
\label{tab:tokenizer_performance}
\begin{tabular}{lcccccccc}
\toprule
Tokenizer & Beat Block & Move Playingcard & Pick Diverse & Move Can & Move Pillbottle & Click & Handover
\\
                            & Hammer $\uparrow$ & Away $\uparrow$ & Bottles $\uparrow$ & Pot $\uparrow$ & Pad $\uparrow$ & Bell $\uparrow$ & Mic $\uparrow$ & \\
\midrule
Binning   & 0.11 & 0.01 & 0.27 & 0.02 & 0.00 & 0.67 & 0.00 \\
FAST      & 0.00 & 0.18 & 0.06 & 0.08 & 0.01 & 0.68 & 0.01 \\
VQ-BET    & 0.10 & 0.27 & 0.12 & 0.13 & 0.06 & 0.64 & 0.00 \\
MethodA(AR)      & 0.25 & 0.47 & 0.14 & 0.65 & 0.35 & 0.83 & 0.18 \\
MethodA(PD)      & 0.12 & 0.49 & 0.05 & 0.44 & 0.46 & 0.80 & 0.18 \\
MethodB(AR)      & 0.16 & 0.60 & 0.22 & 0.62 & 0.36 & 0.90 & 0.22 \\
MethodB(PD)      & 0.29 & 0.63 & 0.21 & 0.47 & 0.41 & 0.89 & 0.28 \\
\bottomrule
\end{tabular}

\vspace{1em}

\begin{tabular}{lccccccc}
\toprule
Tokenizer & Place Container & Handover & Place Phone & Place Burger & Shake & average$\uparrow$ \\
                            & Plate $\uparrow$ & Mic $\uparrow$ & Stand $\uparrow$ & Fries $\downarrow$ & Bottle $\downarrow$ & \\
\midrule
Binning   & 0.00 & 0.85 & 0.02 & 0.01 & 0.92 & 0.24 \\
FAST      & 0.07 & 0.17 & 0.01 & 0.00 & 0.76 & 0.17 \\
VQ-BET    & 0.54 & 0.59 & 0.04 & 0.15 & 0.87 & 0.29\\
MethodA(AR)   & 0.90 & 0.96 & 0.25 & 0.77 & 0.85 & 0.55 \\
MethodA(PD)   & 0.83 & 0.98 & 0.34 & 0.71 & 0.88 & 0.52 \\
MethodB(AR)   & 0.86 & 0.99 & 0.27 & 0.59 & 0.96 & 0.56 \\
MethodB(PD)   & 0.85 & 0.95 & 0.37 & 0.53 & 0.89 & 0.56 \\
\bottomrule
\end{tabular}
\end{table}
\subsubsection{Results Analysis}
We believe such simulation results are interpretable. Introducing state information into the action tokenizer helps the tokenizer perform more precise decoding, thus leading to better experimental performance. The reason why Method B achieves better results than Method A is as follows: In Method A, the decoding process of VQ-VAE\cite{van2017neural} is from discrete codebook tokens to continuous action values, which is essentially limited by the codebook capacity—i.e., there is a one-to-one correspondence between the output tokens in the VLA and the corresponding actions. In contrast, the advantage of Method B lies in extending the decoding process of VQ-VAE\cite{van2017neural} by multiplying the discrete codebook tokens with the action scaling scales provided by the state information. Since the scales obtained from the state network vary with different states, the same token can produce multiple actions. This can be understood as: the tokens in Method B represent only a class of actions, while the state network provides the scale for that class of actions. Therefore, this approach can be extended to a continuous action space, resulting in better performance.
\subsection{Ablation Study}
To validate the effectiveness of our proposed method, we conduct the following ablation studies: (1) verifying whether the introduction of state improves the decoding accuracy of the action tokenizer; (2) evaluating the generalization capability of SA-VLA.
\subsubsection{Ablation Study on the Role of State}
To fully demonstrate the advantage of incorporating state information into the action tokenizer for action decoding, we conducted several comparative experiments. On the aforementioned 12 RoboTwin\cite{chen2025robotwin} tasks, we trained a VLA without state information in the action tokenizer and a VLA with state information in the action tokenizer, and evaluated their average success rates under both autoregressive and parallel decoding settings. The experimental results are shown in Table~\ref{tab:state_comparison}.
\begin{table}[htbp]
\centering
\small
\caption{Ablations on the impact of introducing state to action tokenizer}
\begin{tabular}{c|c}
\hline
\textbf{Method} & \textbf{Average}$\uparrow$ \\
\hline
w/o state (PD) & 0.43 \\
w/o state (AR) & 0.51 \\
Method A (PD) & 0.52 \\
Method A (AR) & 0.55 \\
Method B (PD) & 0.56 \\
Method B (AR) & 0.56 \\
\hline
\end{tabular}
\label{tab:state_comparison}
\end{table}
\subsubsection{Ablation study on model generalization}
To investigate the generalization capability of SA-VLA, we constructed four task settings: \textit{shake\_bottle\_horizontally} and \textit{place\_empty\_cup} in a clean environment, and \textit{handover\_mic} and \textit{place\_container\_plate} in a random environment. The former two tasks are designed to evaluate the model's performance on previously unseen tasks, whereas the latter two tasks aim to assess the model's performance in unseen scenarios. The corresponding experimental results are presented in the Appendix~\ref{sec:Ablations on model generalization}.
\subsubsection{Ablation study on tokenization granularity}
To demonstrate the fine-grained action tokenization capability of our method, we selected four similar action timesteps from four task scenarios on the Beat Block Hammer task in RoboTwin. In tokenizers without state information, these four actions were mapped to identical indices. In contrast, with state-aware tokenization, our method can distinguish these four actions at the level of cosine similarity differences as fine as 0.001.See Appendix~\ref{sec:token_granularity} for details.
\subsection{Real world experiments}
\subsubsection{Experiment Setup}
In our real-world experiments, we employ the AgileX Cobot Magic mobile platform configured with an Aloha setup comprising four robotic arms. Each arm is an AgileX Piper with six degrees of freedom, equipped with a one-DoF parallel gripper. The platform is also equipped with a RealSense D435 RGB camera that captures real-time RGB images at a resolution of $640 \times 480$ pixels and a frame rate of approximately 30 Hz. To evaluate the zero-shot sim-to-real transfer capability of our VLA models, we directly apply the VLAs trained on RGB images from the RoboTwin simulator to three real-world manipulation tasks: Click Bell, Place Container Plate, and Pick Diverse Bottles. Each task is evaluated with 20 trials.The tasks are shown in Appendix~\ref{sec:real world scenes}.

\subsubsection{Real world results}
To demonstrate better results, we adopt the Method B tokenizer, which performs better in simulation, for our experiments. The experimental results in Table~\ref{tab:tokenizer_performance_on_real_world} show that under the zero-shot setting, our method achieves an average success rate of 33\% across the 3 tasks under autoregressive decoding, which is a 23\% improvement over the binning tokenizer, a 25\% improvement over the FAST tokenizer, and an 18\% improvement over the VQ-BET tokenizer.
\begin{table}[htbp]
\centering
\small
\caption{Tokenizer performance comparison on Real world}
\label{tab:tokenizer_performance_on_real_world}
\begin{tabular}{lccccc}
\toprule
Tokenizer & Click bell$\uparrow$ & Place container & Pick Diverse & Average$\uparrow$\\
          & & plate $\uparrow$ & Bottle $\uparrow$ \\
\midrule
Binning   & 6/20 & 0/20 & 0/20 & 0.1 \\
FAST      & 4/20 & 1/20 & 0/20 & 0.08\\
VQ-BET    & 7/20 & 2/20 & 0/20 & 0.15\\
Ours(PD) & 8/20 & 5/20 & 3/20 & 0.27 \\
Ours(AR) & 10/20 & 7/20 & 3/20 & 0.33\\

\bottomrule
\end{tabular}
\end{table}
\section{Conclusion}
\label{sec:conclusion}

	In our work, we experimented with different ways to incorporate state information into the action tokenizer and selected the best action tokenizer to seamlessly integrate into the VLA. In simulation, this achieves up to a 40\% improvement over previous classic action tokenizer approaches, and attains high success rates in real-world sim-to-real experiments.
\section{Limitations}
\label{sec:Limitations}
\textbf{Scalability:} It has only been validated on small-scale datasets, and its scalability has not been demonstrated. Nowadays, many large-scale datasets\cite{deng2026humannet,o2024open} have emerged.Many current works\cite{generalist2025gen0} introduce large-scale datasets to attempt to validate scaling laws\cite{kaplan2020scaling} in robotics, which is also an important future research direction. In subsequent work, we will expand the training data volume and model parameters within our resource limits to attempt to validate the scaling law\cite{kaplan2020scaling} of our method.

\textbf{Model architecture:} Furthermore, we follow the VQ-VAE\cite{van2017neural} architecture; whether other generative models like diffusion\cite{ho2020denoising} can be used to replace VQ-VAE\cite{van2017neural} remains a promising direction for future research. In image generation\cite{yang2026vaevq} and other domains, diffusion\cite{ho2020denoising} models have demonstrated strong generative capabilities and high fidelity. Whether diffusion\cite{ho2020denoising} can be integrated into discrete action tokenizers to improve generation quality is also a direction worth exploring in our future work.

\textbf{Embodiment selection:} Compared to dexterous hands, robotic arms are less convenient in practical real-world production and daily life scenarios. Therefore, whether the embodiment can be migrated from robotic arms to dexterous hands\cite{zhang2026dexora} while maintaining stable performance is a direction we will explore in the future.
\clearpage


\bibliography{example}  
\newpage
\appendix
\centerline{\Large\bfseries Appendix}
\section{Detailed Training And Evaluation Recipes}
\label{app:Detailed Simulation Evaluation Setups}
\subsection{Settings for Data Collection in Simulator}
We conducted experiments in the RoboTwin simulator using two Piper manipulators separated by a distance of 0.6m as the experimental platform. A total of 12 tasks were collected, with 1,600 demonstrations per task. The data collection configuration is shown in the Table~\ref{app:data_collection}.
\begin{table}[htbp]
\centering
\caption{Settings for Data Collection in Simulator}
\label{app:data_collection}
\begin{tabular}{l|c}
\toprule
\textbf{Parameter} & \textbf{Value} \\
\midrule
Save Frequency & 15 \\
Embodiment & Piper \\
Random Background & True \\
Cluttered Table & False \\
Clean Background Rate & 0.02 \\
Random Head Camera Distance & 0.03 \\
Random Table Height & 0.03 \\
Random Light & True \\
Crazy Random Light Rate & 0.02 \\
Head Camera Type & D435 \\
\bottomrule
\end{tabular}
\end{table}
\subsection{Dataset Statistics and Task Metrics}
Table~\ref{app:Dataset Statistics and Task Metrics} summarizes the task names, average trajectory lengths, and a selected instruction for each of the 12 tasks in the simulation.Appendix~\ref{sec:simulation tasks} illustrates some of the tasks and scenarios used in the simulation environment.
\begin{table}[ht]
\centering
\scriptsize
\caption{Dataset Statistics and Task Metrics}
\label{app:Dataset Statistics and Task Metrics}
\renewcommand{\arraystretch}{1.2} 
\begin{tabular}{l|l|l}
\hline
\textbf{Task Name} & \textbf{Avg. Step Num} & \textbf{Task Example Description} \\ \hline
Beat Block Hammer & 67 & Grab the silver hammer and use it to hit. \\ 
Click Bell & 52 & Press the center top of the blue bell. \\ 
Handover Mic & 134 & Grasp the dark blue microphone and pass it across. \\ 
Move Can Pot & 90 & Lift the smooth sauce can, place it by the gray kitchen pot. \\ 
Move Pillbottle Pad & 88 & Hold the white bottle and position it on the pad. \\ 
Move Playingcard Away & 70 & Slide the box for playing cards off the \textit{table} outward. \\ 
Pick Diverse Bottles & 75 & Grab the plastic bottle, pick up the yellow body bottle. \\ 
Place Mouse Pad & 88 & Grab the dark gray mouse and drop it on the black mat. \\ 
Place Container Plate & 92 & Move the white cup and drop the container onto the round plate. \\ 
Place Phone Stand & 82 & Carry the flat phone to the green phone rack. \\ 
Place Burger Fries & 141 & Move the box for hamburg and the red fries box to the tray. \\ 
Shake Bottle & 133 & Shake the orange bottle after lifting it. \\ \hline
\end{tabular}
\end{table}
\subsection{Training Recipe}
The overall training procedure is divided into two stages. In the first stage, the action tokenizer is trained, while in the second stage, the vision-language-action (VLA) model is trained.We train the model using a total of 19,200 trajectories collected from 12 tasks in Appendix~\ref{app:data_collection}. The detailed training recipe is shown in Table~\ref{app:Training recipes}.
\begin{table}[htbp]
\centering
\caption{Training recipes}
\label{app:Training recipes}
\begin{tabular}{l|c|c}
\hline
\textbf{Config} & \textbf{Stage 1} & \textbf{Stage 2} \\
\hline
batch size & 1024 & 64 \\
training epochs & 200 & 10 \\
optimizer & AdamW & AdamW \\
learning rate & $5 \times 10^{-5}$ & $1 \times 10^{-4}$ \\
learning rate schedule & cosine & cosine \\
warmup epochs & 0 & 0 \\
commitment coefficient & 1.0 & 1000.0 \\
adversarial coefficient & 0.1 & 1.0 \\
random horizontal flip & true & false \\
\hline
\end{tabular}
\end{table}
\subsection{Evaluation recipes}
For the evaluation on the Robotwin task, we conduct experiments on a single RTX 4090 GPU. The detailed evaluation procedure can be found at \url{https://robotwin-platform.github.io/doc/index.html}.
\section{Visualization}
\subsection{Visualization on simulation tasks}
\begin{figure}[htbp]
    \centering
    \includegraphics[width=0.6\textwidth]{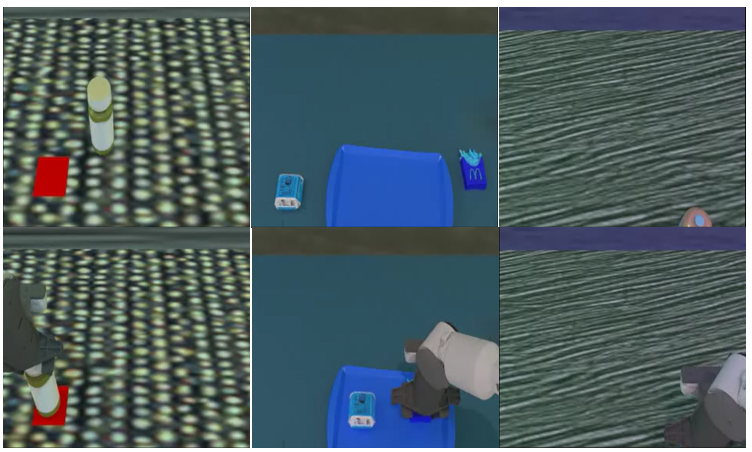}
    \caption{Some of the simulation tasks}
    \label{fig:robotwin任务}
\end{figure}

We selected three task scenarios from RoboTwin, namely \textit{move\_pillbottle\_pad}, \textit{place\_burger\_fries}, and \textit{click\_bell}, and took a partial segment of each scenario, as shown in the Figure~\ref{fig:robotwin任务}.
\label{sec:simulation tasks}
\subsection{Visualization on real world scenes}
\label{sec:real world scenes}
\begin{figure}[htbp]
    \centering
    \includegraphics[width=0.6\textwidth]{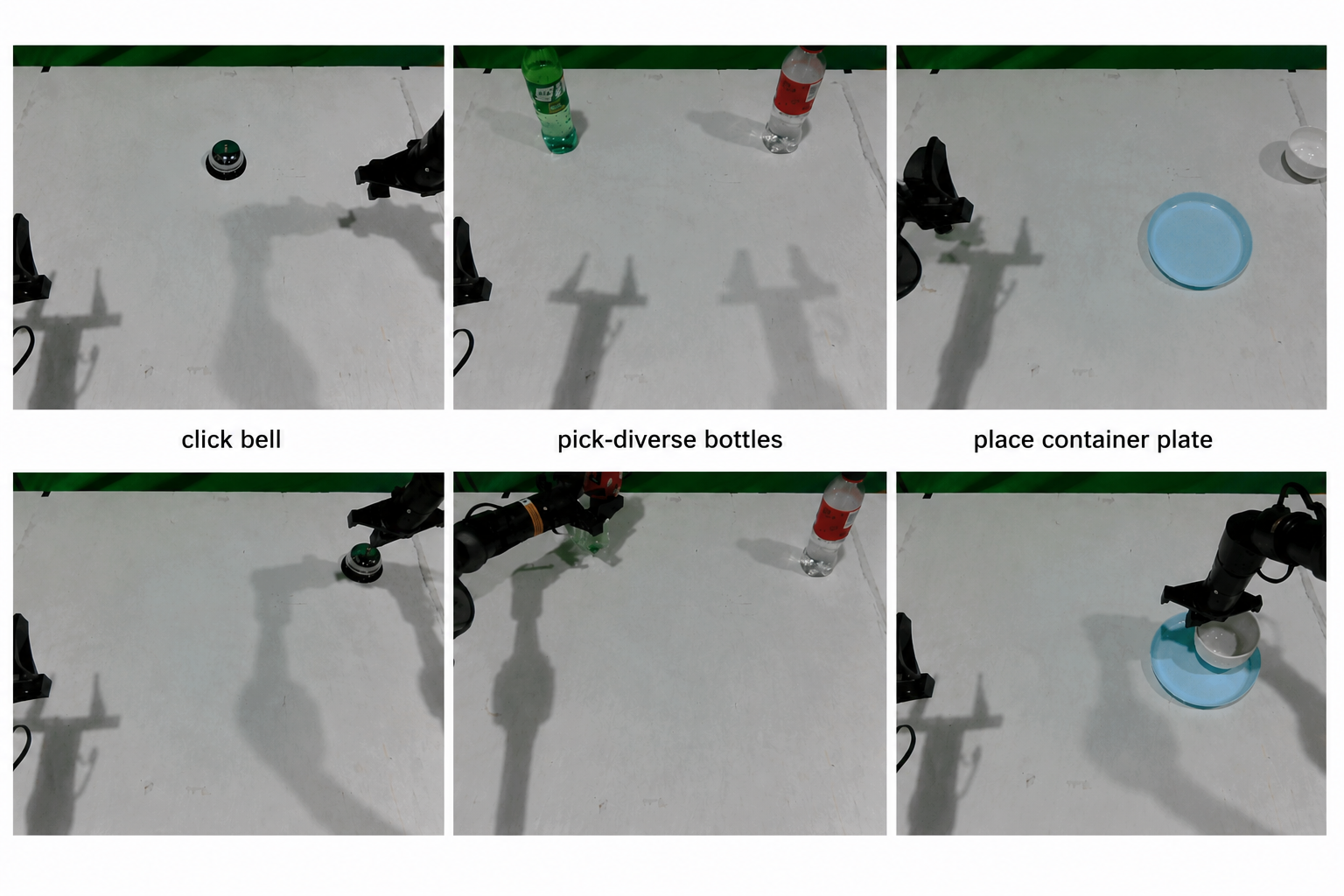}
    \caption{Three task scenarios}
    \label{fig:Three task scenarios}
\end{figure}

In the real-robot experiments, we selected three tasks, namely \textit{Click Bell}, \textit{Place Container Plate}, and \textit{Pick Diverse Bottles}, and took a partial segment of each scenario, as shown in the Figure~\ref{fig:Three task scenarios}.
\section{Ablation study results}
\subsection{Ablations on model generalization}
We selected a total of four tasks for task generalization and scene generalization, and took a partial segment of each scenario, as shown in the Table~\ref{tab:generalization_results}.
\label{sec:Ablations on model generalization}
\begin{table}[htbp]
\centering
\caption{Performance on generalization tasks}
\label{tab:generalization_results}
\begin{tabular}{lc}
\toprule
Task Name & SR \\
\midrule
Shake Bottle Horizontally & 0.95 \\
Place Empty Cup & 0.25 \\
Handover Mic & 0.10 \\
Place Container Plate & 0.16 \\
\bottomrule
\end{tabular}
\end{table}
\subsection{Ablations on tokenization granularity}
\begin{figure}[htbp]
    \centering
    \includegraphics[width=0.6\textwidth]{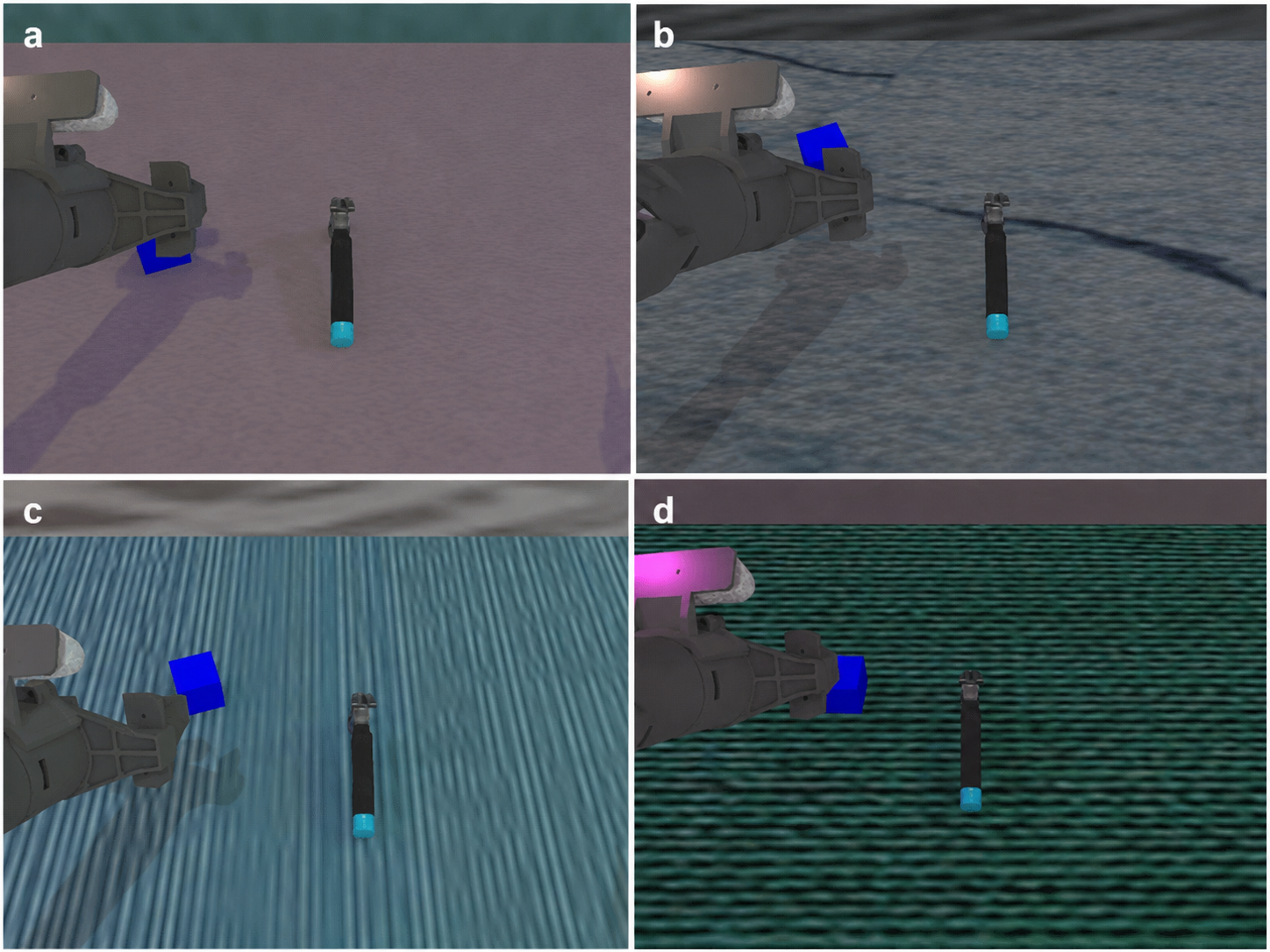}
    \caption{Visualization on tokenization granularity}
    \label{fig:tokenization granularity}
\end{figure}

We observed examples where different actions were mapped to identical tokens by the state-agnostic tokenizer, whereas the state-aware tokenizer mapped them to distinct tokens, as illustrated in Figure~\ref{fig:tokenization granularity} and Table~\ref{table:Similarity comparison between four actions and the first action}. Furthermore, we quantified the changes in reconstruction loss and codebook utilization after incorporating state information, as shown in Figure~\ref{fig:action_tokenizer}. These results demonstrate that incorporating state information enables finer-grained tokenization.
\label{sec:token_granularity}
\begin{table}[htbp]
\centering
\caption{Similarity comparison between four actions and the first action}
\begin{tabular}{c c c c}
\hline
Label & Action & Data Shape & Cosine Similarity to Action 1 \\
\hline
a & Action 1 & $8 \times 7$ & 1.000000\\
b & Action 2 & $8 \times 7$ & 0.999984\\
c & Action 3 & $8 \times 7$ & 0.998856\\
d & Action 4 & $8 \times 7$ & 0.999998\\
\hline
\label{table:Similarity comparison between four actions and the first action}
\end{tabular}
\end{table}
\end{document}